\documentclass{article}

\usepackage{PRIMEarxiv}

\usepackage[utf8]{inputenc} 
\usepackage[T1]{fontenc}    
\usepackage{hyperref}       
\usepackage{url}            
\usepackage{booktabs}       
\usepackage{amsfonts}       
\usepackage{nicefrac}       
\usepackage{microtype}      
\usepackage{lipsum}
\usepackage{fancyhdr}       
\usepackage{graphicx}       
\graphicspath{{media/}}     
\usepackage{graphicx}
\usepackage{booktabs}
\usepackage{multirow}
\usepackage{wrapfig}
\usepackage{floatrow}

\newfloatcommand{figurebox}{figure}[\nocapbeside][\dimexpr(\textwidth-\columnsep)/2\relax]
\newfloatcommand{tablebox}{table}[\nocapbeside][\dimexpr(\textwidth-\columnsep)/2\relax]

\usepackage{amssymb}
\usepackage{graphicx}
\usepackage{array}
\newcolumntype{C}[1]{>{\centering\arraybackslash}p{#1}}
\newcolumntype{L}[1]{>{\arraybackslash}p{#1}}
\newcolumntype{R}[1]{>{\raggedleft\arraybackslash}p{#1}}

\usepackage{algorithm}
\usepackage{algorithmic}
\usepackage{amsmath}

\usepackage{tabularx} 
\usepackage{booktabs} 

\usepackage[accsupp]{axessibility}
\title{DVN-SLAM: Dynamic Visual Neural SLAM Based on Local-Global Encoding
}

\author{
  Wenhua~Wu \\
  Shanghai Jiao Tong University  \\
   \And
  Guangming~Wang \\
  University of Cambridge \\
  \And
  Ting~Deng \\
   Imperial College London\\
  \And
  Sebastian~Aegidius \\
  University College London \\
  \And
  Stuart~Shanks \\
  University College London \\
  \And
  Valerio~Modugno\\
  University College London \\
  \And
  Dimitrios~Kanoulas\\
  University College London \\
  \And
  Hesheng~Wang \thanks{Corresponding Author.}\\
  Shanghai Jiao Tong University \\
}

\begin{document}
\maketitle

\begin{abstract}

Recent research on Simultaneous Localization and Mapping (SLAM) based on implicit representation has shown promising results in indoor environments. However, there are still some challenges: the limited scene representation capability of implicit encodings, the uncertainty in the rendering process from implicit representations, and the disruption of consistency by dynamic objects. To address these challenges, we propose a real-time dynamic visual SLAM system based on local-global fusion neural implicit representation, named DVN-SLAM. To improve the scene representation capability, we introduce a local-global fusion neural implicit representation that enables the construction of an implicit map while considering both global structure and local details. To tackle uncertainties arising from the rendering process, we design an information concentration loss for optimization, aiming to concentrate scene information on object surfaces. The proposed DVN-SLAM achieves competitive performance in localization and mapping across multiple datasets. More importantly, DVN-SLAM demonstrates robustness in dynamic scenes, a trait that sets it apart from other NeRF-based methods.

\end{abstract}    
\section{Introduction}
\label{sec:intro}

\begin{figure*}[t] 
\center{\includegraphics[width=1.0\textwidth]{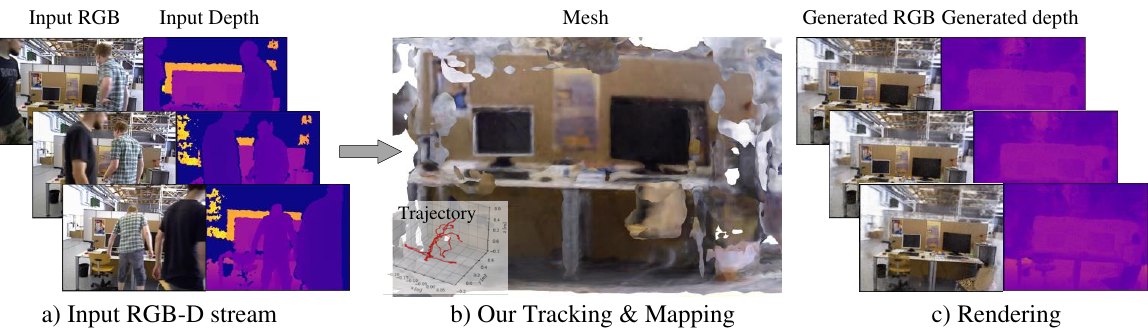}} 
\caption{ We propose DVN-SLAM, a real-time dynamic robust dense visual SLAM system based on local-global fusion neural implicit representation. 
        Compared to current NeRF-based SLAM methods, such as iMAP \cite{sucar2021imap}, NICE-SLAM \cite{zhu2022nice}, Vox-Fusion \cite{yang2022vox}, ESLAM\cite{johari2023eslam} and Co-SLAM \cite{wang2023co}, DVN-SLAM not only achieves competitive performance in static scenes, but also remains effective in high-dynamic scenes. Figure a) is the input dynamic scene video stream. Figure b) showcases the localization and mapping results of DVN-SLAM. Figure c) illustrates the rendering visualization during the mapping process, demonstrating the successful removal of dynamic humans and background completion.
        }
\label{fig:teaser}
\end{figure*}

Dense visual Simultaneous Localization and Mapping (SLAM) is the foundation of perception, navigation, and planning that finds wide applications in areas such as autonomous driving, mobile robotics, and virtual reality. The goal is to enable autonomous systems to accurately estimate their state for localization and synchronously build environmental maps.

Traditional visual SLAM methods primarily focus on localization accuracy and utilize explicit geometric primitives such as surfaces, voxels, and point clouds to represent environmental information~\cite{muglikar2020voxel, gee2008discovering}. While these explicit representations are straightforward, they often suffer from limited accuracy and holes. In recent years, the rapid development of Neural Radiance Fields (NeRF) has demonstrated their tremendous potential in scene modeling~\cite{mildenhall2021nerf, barron2021mip, pumarola2021d, martin2021nerf, yu2022monosdf}. NeRF employs implicit encoding and Multi-Layer Perceptron (MLP) to map positions to spatial information, enabling dense and flexible map construction.

Despite the impressive performance of NeRF-based SLAM in static indoor scenes, there are still several open challenges. Firstly, the scene modeling capability of neural implicit representation is limited. iMAP~\cite{sucar2021imap} utilizes a position-encoded neural implicit representation, and achieves global consistency in scene modeling, but suffers from overly smooth local details and forgetfulness as the scene size increases. On the other hand, methods such as NICE-SLAM~\cite{zhu2022nice} and ESLAM~\cite{johari2023eslam} that are based on feature grids or planes, provide precise modeling of local scene details but exhibit a significant decrease in global representation and predictive ability. Co-SLAM~\cite{wang2023co} combines sparse hash encoding and coordinate encoding, but the joint is simplistic and does not fully exploit their respective advantages. In this paper, we present a novel local-global fusion neural implicit representation by employing both attention-based feature fusion and result fusion. The proposed neural implicit representation effectively leverages the advantages of the continuous neural radiance field for global representation and the discrete grid for local representation, enabling the modeling of both local details and global structure.

Furthermore, the optimization of neural radiance fields relies on supervised losses between rendered images and observed images~\cite{mildenhall2021nerf}. However, the process of volume rendering introduces certain uncertainties, where different information distributions along the same viewing ray can yield the same rendering result upon integration. This uncertainty implies that even with minimal rendering errors, the distribution of scene information may not be accurate. To address this issue, we incorporate information concentration loss based on rendering variance, aiming to concentrate information along the observed rays. This aligns with the prior knowledge that color information tends to be concentrated on object surfaces.

Despite the good performance of existing NeRF-based SLAM methods in static indoor scenes, they struggle with dynamic scenes. The motion of objects disrupts the static consistency of the scene, making the pure pose implicit mapping inadequate for modeling. Although some offline methods for dynamic scene rendering have achieved dynamic modeling through the introduction of temporal fields~\cite{pumarola2021d, li2021neural,yan2023nerf}, such modeling approaches often require extensive training and are challenging to apply in SLAM. Furthermore, SLAM focus lies on localization and consistent mapping, and real-time modeling of moving objects is usually unnecessary. 
Thanks to the designed neural implicit representation based on local-global fusion, our approach can maintain stable scene structure modeling in dynamic scenarios. Rapidly moving objects can be automatically ignored, and the background occluded by dynamic objects can be effectively recovered. In contrast, existing NeRF-based SLAM methods such as ESLAM~\cite{johari2023eslam} and Co-SLAM~\cite{wang2023co} that employ simplistic scene representations are prone to scene structure disruption by dynamic objects, resulting in complete failure in localization and mapping.

The main contributions of our work can be summarized as follows:

\begin{itemize}
  \item We propose DVN-SLAM, a real-time dynamic robust dense visual SLAM system. The core of DVN-SLAM is a local-global fusion neural implicit representation, which employs both attention-based feature fusion and result fusion to leverage the advantages of the local discrete grid and the global continuous neural radiance field. 
  \item We consider the uncertainty of the rendering process and design an information concentration loss for optimization. 
  \item We perform extensive evaluations on multiple datasets and real-world scenarios, demonstrating that DVN-SLAM not only achieves competitive performance in static scenes, but also remains effective in high-dynamic scenes while existing methods fail.

\end{itemize}

\section{Related Work}
\label{sec:related_work}

\textbf{Dense Visual SLAM Prior to NeRF.} Visual SLAM systems focus on localizing and mapping by collecting data with a monocular camera. The foundation of dense visual SLAM lies in DTAM~\cite{newcombe2011dtam}, which pioneers real-time tracking and mapping via dense scene representation. In the same period, the KinectFusion method~\cite{newcombe2011kinectfusion} makes notable strides by using ICP algorithms~\cite{besl1992method} and volumetric Truncated Signed Distance Function (TSDF) to achieve accurate and real-time reconstruction of dense surfaces for indoor scenes. Lots of innovative data structures including Surfels~\cite{whelan2015elasticfusion,schops2019surfelmeshing} and Octrees~\cite{vespa2018efficient,xu2019mid}, are proposed to improve scalability and reduce memory. In contrast to these methods which rely on per-frame pose optimization, BAD-SLAM~\cite{schops2019bad} is the first to propose a full Bundle Adjustment (BA) to jointly optimize the keyframes. Recently, numerous deep learning-based SLAM methods~\cite{bloesch2018codeslam, li2020deepslam, koestler2022tandem, peng2020convolutional, teed2021droid} are introduced to improve the precision and robustness of traditional SLAM methods with various learnable parameters.

\noindent\textbf{Neural Implicit Representations.} 
Neural Radiance Field (NeRF)~\cite{mildenhall2021nerf} is a novel neural implicit representation method that utilizes an MLP to model the mapping from spatial locations to the geometry and appearance information of scene, achieving remarkable results through the principle of volume rendering. The initial NeRF~\cite{mildenhall2021nerf} could only model static individual objects, while subsequent methods have studied the implicit representation of challenging scenes involving large scale~\cite{tancik2022block,turki2022mega, turki2023suds}, dynamics~\cite{pumarola2021d, li2021neural,yan2023nerf}, lighting variations~\cite{martin2021nerf,chen2022hallucinated}, and other factors.  Furthermore, some methods investigate how to enhance the learning efficiency of implicit representation~\cite{muller2022instant}.

\noindent\textbf{Neural Implicit SLAM.} Recent work on 3D scene reconstruction implies the advantage of both SLAM and NeRF, achieving an accurate and real-time reconstruction for static indoor scenes~\cite{zhu2022nice}. One classical work is iMAP~\cite{sucar2021imap}, which performs jointly real-time mapping and tracking, plausible filling to the unobserved regions via trained MLP. Subsequently, NICE-SLAM~\cite{zhu2022nice} improves the iMAP~\cite{sucar2021imap} model by introducing a hierarchical feature grid for scene representation and optimizing it using pre-trained decoders. Although NICE-SLAM~\cite{zhu2022nice} achieved higher scalability, accuracy, and robustness compared to iMAP~\cite{sucar2021imap}, it is still limited to novel scenes and lacks reconstruction details. Point-SLAM~\cite{sandstrom2023point} utilizes a point-based neural implicit representation to achieve more precise modeling of scene details. However, its strategy of increasing scene points incurs significant computational overhead, resulting in poor real-time performance.

The most related works to ours are ESLAM~\cite{johari2023eslam} and Co-SLAM~\cite{wang2023co}. Compared to NICE-SLAM~\cite{zhu2022nice}, the main novelties of ESLAM are the use of axis-aligned feature planes and decoding them into a TSDF to present scene reconstructions, resulting in reduced memory growth rate and increased reconstruction efficiency as well as accuracy. This grid-based feature representation enables local fine-grained modeling but at the cost of significantly reduced global predictive capability. Co-SLAM~\cite{wang2023co} integrates sparse hash encoding and coordinates encoding, but it only applies a simple joint of the two without fully exploiting their individual advantages. In contrast, we employ both attention-based fusion and result fusion to fuse feature planes and position encoding, effectively leveraging the advantages of both the local discrete grid and the global continuous neural radiance field. Current methods lack consideration for the uncertainties introduced during the rendering process, we design an information concentration loss for optimization. Additionally, existing methods fail to handle dynamic scenes. While NICE-SLAM~\cite{zhu2022nice} performs simple filtering based on abnormal loss values in the tracking thread, it cannot completely eliminate the influence of dynamic objects. In contrast, the stronger modeling capability of the proposed local-global fusion representations enables our method to remain effective in dynamic scenes.

\begin{figure*}[t] 
\center{\includegraphics[width=1.0\textwidth]{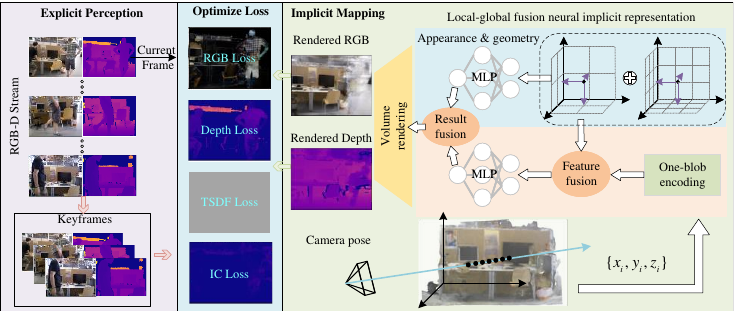}} 
\caption{Overview of DVN-SLAM. The left side of the framework is the explicit perception module, which takes RGB-D video streams as input. The decision of adding a frame as a keyframe is based on the information gain from the image. On the right side of the framework is the local-global fusion neural implicit representation, which establishes a mapping from spatial locations to color and TSDF values. Volume rendering is performed to generate RGB-D images corresponding to the tracking poses. The middle is our loss optimization module, which includes color loss, depth loss, TSDF loss, and information concentration loss.}
\label{framework}
\end{figure*}

\section{Method}\label{sec:method}

The overview of the proposed DVN-SLAM method is shown in Fig.~\ref{framework}. The left side of the framework is the explicit perception module. On the right side of the framework is the local-global fusion neural implicit representation. The middle is our loss optimization module. Each component will be introduced in detail: Sec.~\ref{fusion} introduces the local-global fusion neural implicit representation, while Sec.~\ref{render_loss} shows 
rendering and loss functions.
Sec.~\ref{tracking_mapping} describes the tracking and mapping optimization process.

\subsection{Global-Local Fusion Neural Implicit Representation}
\label{fusion}

\begin{figure}[htb] 
\center{\includegraphics[width=1.0\columnwidth]{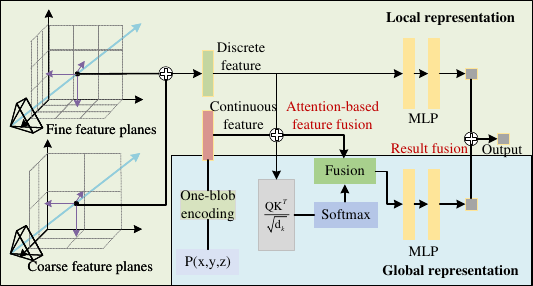}} 
\caption{Local-global fusion neural implicit representation. The blue region represents the global representation, while the green region represents the local representation. We employ two methods, as shown in the red, attention-based feature fusion, and result fusion, to merge the global and local representations, achieving stronger representational capacity and more stable neural implicit representation.}
\label{fusion_mlp}
\end{figure}

The original neural radiance field utilizes positional encoding to map coordinates to high-frequency features and then uses MLP to output color and geometry information~\cite{mildenhall2021nerf}. Benefiting from the continuity of coordinate space and the smoothness of MLP, when trained offline with a sufficient number of multi-view images, it can achieve accurate scene modeling. However, in the context of SLAM systems, images are obtained in real-time as the camera moves, and there are not enough multi-view observations. Moreover, SLAM systems require high real-time performance. This positional-encoding-based implicit representation tends to result in overly smooth scene details and convergence difficulties. Additionally, scene information is implicitly stored in the global MLP, and each optimization step alters the global information. The neural implicit representation based on discrete feature planes is able to store scene information in local features by synchronously optimizing the parametrized features and MLP parameters. The MLP learns how to decode the features. However, this loses the global prediction capability provided by coordinate continuity, which means it cannot predict unobserved regions and fill in the holes.

To address these challenges, we propose a local-global fusion neural implicit representation, shown in Fig.~\ref{fusion_mlp}, which employs both attention-based feature fusion and result fusion, aiming to leverage the advantages of both continuous neural radiance fields and discrete feature planes while minimizing their drawbacks. In our proposed model, we employ the One-blob~\cite{muller2019neural} encoding $f_{o}(p)$ for global representation. For local representation, we use hierarchical axis-aligned feature planes, with coarse and fine geometric feature planes $F^c_g = \{F^c_{g-XY}, F^c_{g-XZ}, F^c_{g-YZ}\}$, $F^f_g =\{F^f_{g-XY}, F^f_{g-XZ}, F^f_{g-YZ}\}$ and color feature planes $F^c_a =\{F^c_{a-XY}, F^c_{a-XZ}, F^c_{a-YZ}\}$, $ F^f_a=\{F^f_{a-XY}, F^f_{a-XZ}, F^f_{a-YZ}\}$. Given the point coordinate $p= (x,y,z)$, we query their components on each plane and sum them to obtain local features, geometric feature $f_g(p)$ and appearance feature $f_{a}(p)$:
\begin{equation}
\underset{i \in \{a,g\}, l \in \{c,f\}}{{f^l_i}(p)} = F^l_{i-XY}(x,y) +F^l_{g-XZ}(x,z) +F^l_{g-YZ}(y,z),
\end{equation}
\begin{equation}
f_g(p) = [f^c_g(p),f^f_g(p)];f_a(p) = [f^c_a(p),f^f_a(p)],
\end{equation}
where, the superscripts $l \in \{c,f\}$  mean coarse and fine levels. The subscripts $i \in \{a,g\}$ mean geometric and appearance. $\{XY, XZ, YZ\}$ are coordinate planes. The operation of $[ \;\cdot\;,\; \cdot\;]$ denotes feature concatenation.

\noindent \textbf{Attention-based feature fusion.} 
We concatenate the global features and local features and use an attention module for fusion:
\begin{equation}
\begin{split}
   f_i^{cat}(p) &= [f_i(p),f_o(p)],\\
   {f_{i}}^{att}(p) &= Softmax(\frac{f_i^{cat}(p) f_i^{cat}(p)^T}{\sqrt{d_i}})f_i^{cat}(p),
\end{split}
\end{equation}
where, $i \in \{a,g\}$ and $d_i$ is the dimension of $f_i^{cat}(p)$. ${f_{i}}^{att}(p)$ is the final fused feature. The attention fusion combines different features of the same point, including the grid feature from local representation and the pose encoding feature from global representation, enabling better fusion of local and global representations than direct concatenation.

\noindent\textbf{Result fusion.} 
We use MLPs to decode the fusion features and local features respectively, and then fuse the results to get the final geometry and color information:
\begin{equation}
c_p = \omega \phi^1_{a}({f_{a}}^{att}(p)) + (1-\omega) \phi^2_{a}({f_{a}}(p)), 
\end{equation}
\begin{equation}
s_p = \omega \phi^1_{g}({f_{g}}^{att}(p)) + (1-\omega) \phi^2_g({f_{g}}(p)), 
\end{equation}
where $\{\phi^1_{a}, \phi^2_{a}, \phi^1_{g}, \phi^2_{g}\}$ are feature decoders. $\omega$ is the fusion weight. $c_p$ is the RGB value, and $s_p$ is the TSDF value.

It is important to note that although Co-SLAM also uses both coordinate and hash encodings, it simply concatenates them together, and the color and geometry predictions are coupled. In contrast, our approach combines discrete local features and continuous coordinate encoding through attention-based feature fusion and result fusion, achieving a better balance between local scene details and hole-filling. More importantly, this fusion design makes our method insensitive to dynamic objects and able to maintain the scene structure modeling while other methods are completely ineffective.

\subsection{Rendering and Loss Functions}
\label{render_loss}

\noindent \textbf{Color and depth rendering.} 
We follow the same sampling strategy as ESLAM~\cite{johari2023eslam}, where $N$ points are sampled along each ray.
The implicit representation model from the Sec.~\ref{fusion} predicts the color $\{c_i\}$ and TSDF $\{s_i\}$ for each sampled point. We employ the following transformation function to convert the TSDF into weights $\{w_i\}$:
\begin{equation}
w_i = \sigma(\frac{s_i}{tr})\sigma(-\frac{s_i}{tr}),
\end{equation}
where $tr$ is the truncation distance and $\sigma$ represents the Sigmoid function. This transformation function ensures that the weight is maximum at the location where the TSDF equals 0, indicating the surface of the object.

Then, the color value and depth value of the pixel are obtained by weighted summing the color and depth of the sampled points on the ray:
\begin{equation}
\hat{c} = \frac{1}{\sum_{i =1}^{N}w_i}  \sum_{i= 1}^{N} w_i c_i,
\hat{d} = \frac{1}{\sum_{i =1}^{N}w_i}  \sum_{i= 1}^{N} w_i d_i.
\end{equation}

\noindent \textbf{Color and depth loss.} In order to reduce the computational burden, we randomly sample $M$ points for each image to render, and then calculate the supervision loss of rendering and observing RGB-D images:
\begin{equation}
\begin{split}
    L_{rgb} = \frac{1}{M}  \sum_{m= 1}^{M}(\hat{c}_m - c_m)^2, \\
L_{d} = \frac{1}{M_d}  \sum_{m= 1}^{M_d}(\hat{d}_m - d_m)^2,
\end{split}
\end{equation}
where $M_d$ is the number of pixels where the depth value is valid. $M_d \leq M$, because there are holes in the depth map.

\begin{table*}[t]
\centering
\resizebox{\textwidth}{!}{
\begin{tabular}{c|cccc|cc}
\toprule
\multirow{2}{*}{Methods} & \multicolumn{4}{c|}{Reconstruction}                               & \multicolumn{2}{c}{Localization}    \\
& $Depth \: L1[cm] \downarrow $ & $Acc.[cm]\downarrow $ & $Comp.\:[cm]  \downarrow$ & $Comp.Ratio(\%) \uparrow $ & $ATE \: Mean[cm] \downarrow$ & $ATE \: RMSE[cm]\downarrow$ \\                      
\midrule
\multicolumn{1}{l|}{iMAP* \cite{sucar2021imap}} & 4.64 & 3.62 & 4.93 & 80.51 & 2.59 & 3.42 \\
\multicolumn{1}{l|}{NICE-SLAM \cite{zhu2022nice}} & 1.90 & 2.37 & 2.64 & 91.13 & 1.56 & 2.05 \\
\multicolumn{1}{l|}{Vox-Fusion \cite{yang2022vox}} & 2.91 & \textbf{1.88} & 2.56 & 90.93 & 1.02 & 1.47 \\
\multicolumn{1}{l|}{Co-SLAM \cite{wang2023co}} & 1.51 & 2.10 & 2.08 & 93.44 & 0.94 & 1.06\\
\multicolumn{1}{l|}{ESLAM \cite{johari2023eslam}} & \underline{0.94} & 2.18 & \underline{1.75} & \textbf{96.46} & \underline{0.52} & \underline{0.64} \\
DVN-SLAM (Ours) &\textbf{0.75} & \underline{2.09} & \textbf{1.70} & \underline{96.43} & \textbf{0.45} & \textbf{0.53} \\
\bottomrule
\end{tabular}}
\caption{ Quantitative comparison of DVN-SLAM with existing nerf-based dense V-SLAM in terms of reconstruction and localization accuracy on the Replica dataset \cite{straub2019replica}. The reconstruction metrics include L1 loss (cm) between rendered images from 1000 randomly sampled camera poses and ground truth depth maps, reconstruction accuracy (cm), completion accuracy (cm), and completion rate (\%). The localization metric is measured by mean and RMSE of Absolute Trajectory Error (ATE) (cm). The results are reported as the average over 5 runs on 8 scenes of the Replica dataset \cite{straub2019replica}. Our approach demonstrates significant improvements over previous works. Note that iMAP* denotes the iMAP \cite{sucar2021imap} reimplementation released by the NICE-SLAM \cite{zhu2022nice} authors. For a detailed evaluation of each scene, please refer to the appendix.}
\label{replica_tab}
\end{table*}

\noindent \textbf{TSDF loss.}
In addition to supervising the rendering results, the geometric representation using TSDF allows for supervision at any point along the ray. For points within the truncation region $P_{in}$, we use the difference between the pixel depth value and the sampled point depth as the ground truth for the TSDF value, providing supervision:
\begin{equation}
L_{sdf}(P_{in}) = \frac{1}{M_d} \sum_{m=1}\frac{1}{|P_{in}|} \sum_{p \in P_{in} }(s_p - (d-d_p))^2.
\end{equation}

Follow ESLAM~\cite{johari2023eslam}, $P_{in}$ is divided into $P^n_{in}$ and $P^f_{in}$, where $P^n_{in} = \{p \mid |d_p-d|<0.4tr\}$, $P^f_{in} =P_{in} - P^n_{in} $. 
Therefore, the losses in the two areas are calculated as $L^n_{sdf}(P^n_{in})$ and $L^f_{sdf}(P^f_{in})$ respectively.

For points outside the truncation region $P_{out}$, we employ a free-space loss that encourages the TSDF values to approach the truncation distance as follows:
\begin{equation}
L_{fs} = \frac{1}{M_d} \sum_{m=1} \frac{1}{|P_{out}|} \sum_{p \in P_{out} }(s_p -tr)^2.
\end{equation}

\begin{figure}[t] 
\center{\includegraphics[width=0.32\textwidth]{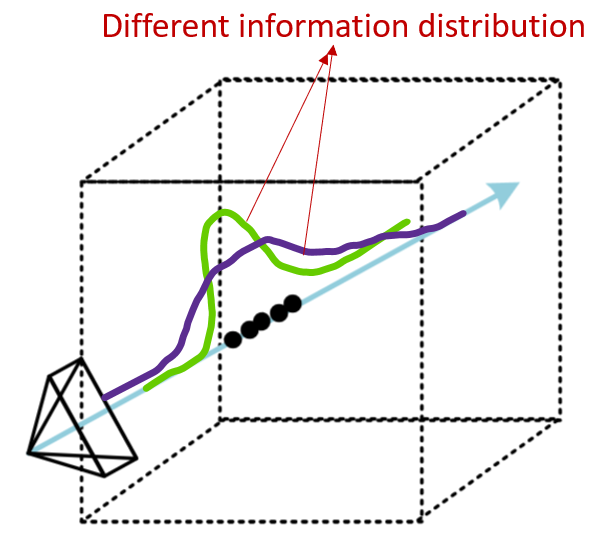}} 
\caption{Illustration of Volume Rendering. Different information distributions along the same ray may yield the same rendering result. This introduces uncertainty in the distribution of scene information even when the rendering result is determined.
}
\label{render}
\end{figure}

\noindent \textbf{Information concentration loss.} 
As shown in Fig.~\ref{render}, the volumetric rendering process is inherently uncertain, as there is a possibility that different information distributions along a ray can yield the same rendered result. Due to this uncertainty, even with small rendering loss, there is no guarantee that the learned scene information is accurate. To address this problem, we propose a depth-variance-based information concentration loss. Specifically, we compute the depth variance along the ray and aim to minimize it.
\begin{equation}
\hat{d}_{var} = \frac{1}{\sum_{i =1}^{N}w_i}  \sum_{i= 1}^{N} w_i (d_i - \hat{d})^2.
\end{equation}
\begin{equation}
L_{ic} = \frac{1}{M_d}  \sum_{m= 1}^{M_d}\hat{d}_{var}.
\end{equation}

By optimizing this loss, we encourage the distribution of scene information to concentrate on the surface of scenes.

The total loss function is defined as:
\begin{equation}
\begin{split}
L_{total} = \lambda_{rgb} L_{rgb} +\lambda_{d} L_{d} +\lambda^n_{sdf} L^n_{sdf}
\lambda^f_{sdf} L^f_{sdf}+\lambda_{fs} L_{fs}  +\lambda_{ic} L_{ic},
\end{split}
\end{equation}
where $\{\lambda_{rgb}, \lambda_{d}, \lambda^n_{sdf},\lambda^f_{sdf}, \lambda_{fs}, \lambda_{ic}\}$ are loss weights.

\subsection{Mapping and Tracking}
\label{tracking_mapping}

Given an initial pose and the first frame, we render images from this pose and minimize color and depth loss, TSDF loss, and information concentration loss to initialize the neural implicit representation. 

\noindent \textbf{Tracking.} For each observed frame, using the previous frame pose as the source pose, we do rendering and calculate losses, and iteratively optimize to obtain the localization result.

\noindent \textbf{Mapping.} Following~\cite{johari2023eslam}, we apply a keyframe selection every $k$ frames and perform map optimization. We select $W$ frames including the current frame, the previous two keyframes, and $W - 3$ frames randomly selected from the keyframe list. We do rendering and calculate losses, and iteratively optimize the feature planes, fusion network, and corresponding camera poses. 

\begin{figure*}[t] 
\center{\includegraphics[width=1.0\textwidth]{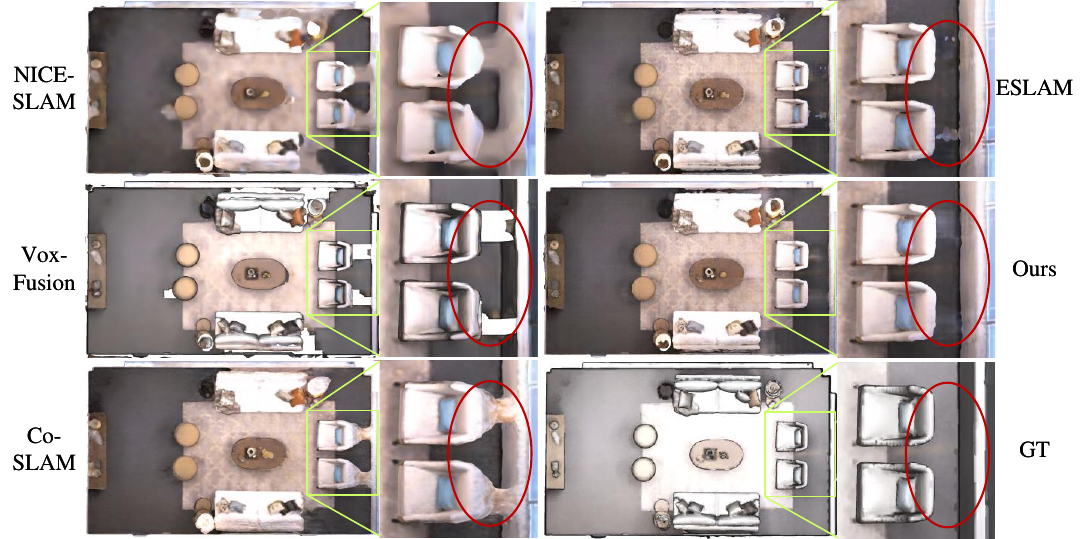}} 
\caption{Reconstruction results of room0. The chair region is magnified to highlight. Compared to other methods, DVN-SLAM can achieve accurate modeling of local details and global structure, providing more reasonable predictions for unobserved regions, such as the backside of the chair shown in the figure.
}
\label{replica_result}
\end{figure*}

\section{Experiment}
\label{sec:experiment}
We conduct experiments to demonstrate the superior localization and mapping accuracy, as well as dynamic robustness, of the DVN-SLAM compared to other methods.

\subsection{Experiment Setup}
\textbf{Datasets.}
Following previous work, we conduct experimental evaluations on two standard 3D datasets: the Replica dataset~\cite{straub2019replica}
and the TUM\_RGBD dataset~\cite{sturm2012benchmark}. In addition to the static scenes commonly used in previous work, we do experiments in challenging dynamic scenes from the TUM\_RGBD dataset~\cite{sturm2012benchmark}. 

\noindent\textbf{Implementation details.}
The resolution of coarse and fine feature planes is $\{24$ cm, $6$ cm$\}$ for geometry and $\{24$cm, $3$ cm$\}$ for appearance. The feature dimension is $24$. The MLPs in the neural implicit representation consist of two layers with 16 channels, using the ReLU activation function. The truncation distance $tr=6$ cm. In the tracking process, we sample $2000$ pixels per image. The loss weights are set as $\lambda_{rgb} = 5, \lambda_{d} = 1, \lambda^n_{sdf} = 200, \lambda^f_{sdf} = 50,\lambda_{fs}=5, \lambda_{ic}=0.$ In the mapping process, we sample $4000$ pixels per image. The loss weights are set as $\lambda_{rgb} = 5, \lambda_{d} = 0.1, \lambda^n_{sdf} = 200, \lambda^f_{sdf} = 10,\lambda_{fs}=5, \lambda_{ic}=0.05$. More experimental details can be found in the supplementary materials. All experiments are conducted on a server equipped with an NVIDIA A100 GPU.


\noindent\textbf{Metrics.}
We evaluate the reconstruction quality using depth L1 (cm), accuracy (cm), completeness (cm), and completion rate (\%) thresholds set to $5$ cm. Following the evaluation methodology of Co-SLAM~\cite{wang2023co}, we remove any unobserved regions outside the camera frustum and perform an additional grid culling to remove noise points inside the camera frustum but outside the target scene. For tracking evaluation, we utilize the Average Translation Error (ATE) mean and Root Mean Square Error (RMSE) in centimeters.

\begin{table*}[t]
\centering
\resizebox{\textwidth}{!}{
\begin{tabular}{c|ccc|ccc}
\toprule
\multirow{2}{*}{Methods} & &Static & & &Dynamic & \\
& fr1/desk&  fr2/xyz  &fr3/office &fr3/walking-halfshere & fr3/walking-xyz & fr3/walking-static  \\                   
\midrule
\multicolumn{1}{l|}{iMAP* \cite{sucar2021imap}}     &7.20& 2.10& 9.00 & 9099.5 & 127.97 & 1434.9 \\
\multicolumn{1}{l|}{NICE-SLAM \cite{zhu2022nice}} &2.98&1.62&4.05& 692.27 & 120.17 & \underline{67.94} \\
\multicolumn{1}{l|}{Co-SLAM \cite{wang2023co}}   &2.98&1.83&\textbf{2.76}& 144.86 & 98.48 & 88.88 \\
\multicolumn{1}{l|}{ESLAM \cite{johari2023eslam}}&\underline{2.54}& \textbf{1.14}  & 2.81 & \underline{79.24} &  \underline{96.50}  &70.22   \\
\multicolumn{1}{l|}{Point-SLAM~\cite{sandstrom2023point}}   &4.34&1.31& 3.48 &  124.05 & 856.39  & - \\
\multicolumn{1}{l|}{DVN-SLAM (Ours) } &\textbf{2.29}& \underline{1.25}& \underline{2.78} &\textbf{51.77} &\textbf{17.68} & \textbf{2.05} \\

\bottomrule
\end{tabular}}
\caption{ATE RSME (cm) results on TUM-RGBD dataset. Compared to other Nerf-based SLAM methods, our approach achieves remarkable localization results in static scenes and dynamic scenes. Particularly, other methods fail in dynamic scenes, while our method continues to perform well. Note that iMAP* denotes the iMAP \cite{sucar2021imap} reimplementation released by the NICE-SLAM \cite{zhu2022nice} authors. "-" for Point-SLAM~\cite{sandstrom2023point} in fr3/walking-static means a complete failure.}
\label{tum_tab}
\end{table*}

\begin{figure*}[h] 
\center{\includegraphics[width=1.0\textwidth]{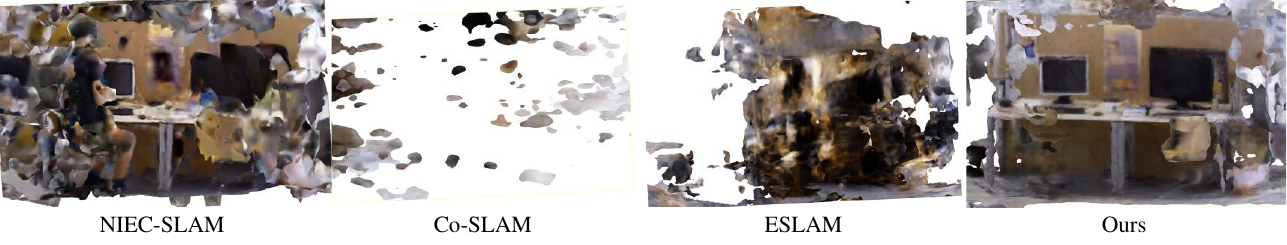} 
\caption{Reconstruction results on TUM-RGBD. In dynamic scenes, our method effectively ignores dynamic people and successfully reconstructs the static background, while other methods completely fail. 
}
\label{tum_result}
}
\end{figure*}
\subsection{Experiment Results}
\textbf{Evaluation on Replica~\cite{straub2019replica}.}
We conducted experiments on $8$ scenes from the Replica dataset \cite{straub2019replica}. The average results
over $5$ runs on $8$ scenes are presented in Tab.~\ref{replica_tab}. Compared to other Nerf-based SLAM algorithms, our approach achieves state-of-the-art performance in terms of localization accuracy, Depth L1, and completeness (Comp.), while ranking second in terms of accuracy (Acc) and completeness ratio (Comp. Ratio). Due to the adoption of an Octree for scene representation in Vox-Fusion \cite{yang2022vox}, the reconstructed results tend to have more holes, resulting in a high accuracy (Acc) but a low completeness ratio. 

\begin{figure}[t] 
\center{\includegraphics[width=\columnwidth]{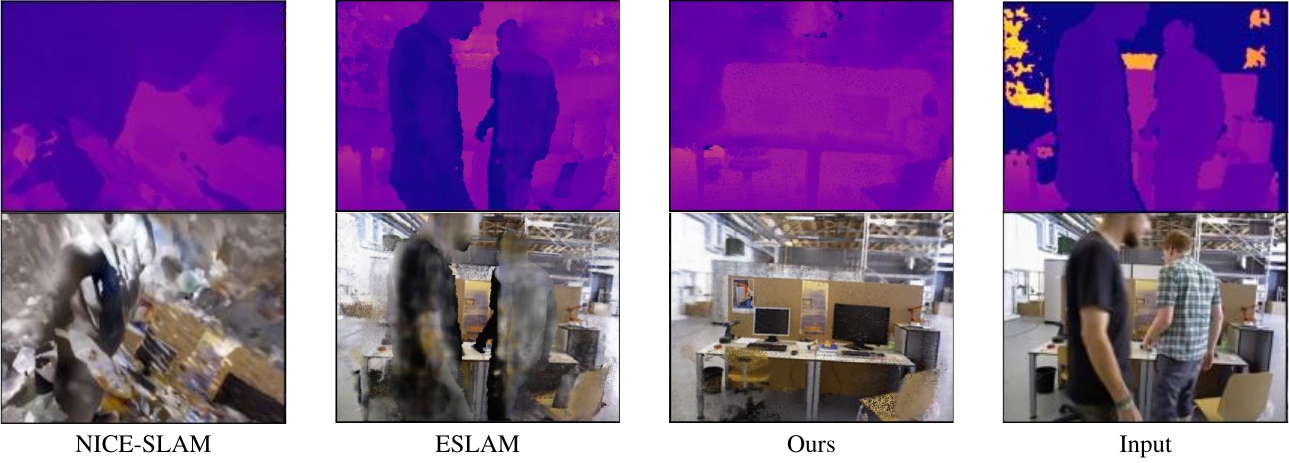} 
\caption{Rendering results in dynamic scenes. Our method not only produces clearer rendering in static regions but also effectively filters out dynamic objects and fills in the corresponding background. In contrast, other methods produce blurry renderings and retain dynamic occlusions.
}
\label{tum_result_rgbd}
}
\end{figure}

Fig.~\ref{replica_result} illustrates a visualization comparison of the reconstruction results between DVN-SLAM and others. It can be observed that our method not only captures finer details but also demonstrates superior predictive capabilities for unobserved regions. 
The chair region is magnified to highlight. In this sequence, the backside of the chair is not observed from any viewpoint. Vox-Fusion \cite{yang2022vox} utilizes an octree structure for scene representation, resulting in significant holes in the reconstruction. ESLAM \cite{johari2023eslam} employs feature-plane-based scene representation, lacking global prediction capability, leading to holes on the front side of the chair and noticeable protrusions on the backside. Co-SLAM \cite{wang2023co} simply concatenates sparse hash encoding and coordinate encoding, resulting in excessive filling of unobserved areas. In contrast, DVN-SLAM predicts a smooth wall and floor surface. This ability to achieve fine-grained reconstruction and global prediction stems from the proposed local-global fusion neural implicit representation.


\noindent\textbf{Evaluation on TUM\_RGBD \cite{sturm2012benchmark}}
Tab.~\ref{tum_tab} presents the evaluation results in static scenes and dynamic scenes of the TUM-RGBD dataset. Compared to other NeRF-based methods, our approach achieves remarkable performance consistently. Particularly in dynamic scenes, our method remains effective. The poor performance in fr3/walking-halfshere is due to the moving people being too close to the camera, while other methods perform even worse.

\begin{wrapfigure}{l}[0cm]{0pt}
\includegraphics[width=0.5\columnwidth]{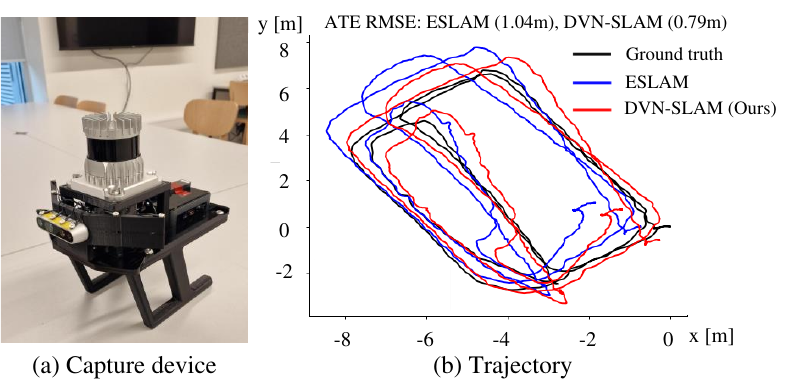} 
\caption{Capture device and localization 
result in real-world scenes. Our method proves to be effective when applied in practical scenarios.
}
\vspace{-10pt}
\label{rpl_result}
\end{wrapfigure}

Fig.~\ref{tum_result} showcases the reconstruction results of the fr3/walking-xyz. Our method successfully ignores fast-moving individuals and reconstructs the static background, while other methods completely fail. Fig.~\ref{tum_result_rgbd} illustrates the rendering visualization during the mapping process. Our method produces clearer rendering in static regions, and it effectively filters out dynamic objects and fills in occluded backgrounds. This is due to the designed local-global fusion neural implicit representation, which makes our method insensitive to dynamic objects and can maintain the scene structure modeling.

\noindent\textbf{Evaluation on real-world scene}. We captured a sequence using a handheld RGB-D camera, as shown in Fig.~\ref{rpl_result} (a), in our laboratory to assess the performance of our method in real-world applications. The sequence consists of 6677 pairs of RGB-D images with a resolution of $1280\times720$. The capture frequency was $30$Hz. Fig.~\ref{rpl_result} (b) showcases the localization 
results. It can be observed that DVN-SLAM exhibits superior accuracy compared to ESLAM~\cite{johari2023eslam} in real-world scenes, thereby validating its practicability.

\subsection{Ablation Study}
\begin{figure}[h] 
\center{\includegraphics[width=\columnwidth]{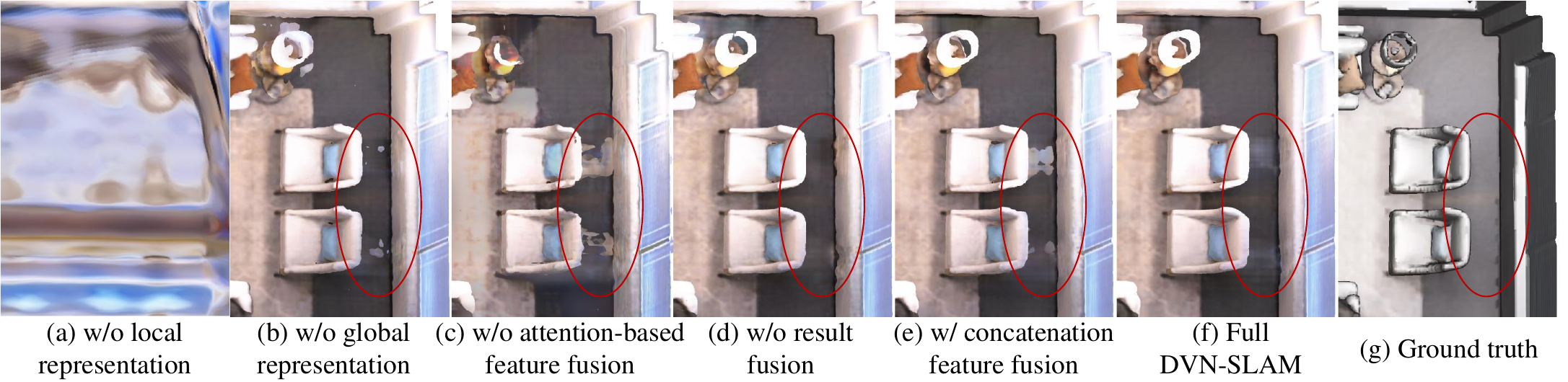}
\caption{Visualization of reconstruction in ablation experiments. The back of the chairs is highlighted with red circles, demonstrating the effectiveness of the local-global representation in making reasonable and smooth predictions for unobserved regions.
}
\label{ic_loss_ab}
}
\end{figure}

\noindent \textbf{Effect of local-global fusion neural implicit representation.}
Tab.~\ref{ablation} presents the quantitative evaluations of different neural implicit representations. Our complete model outperforms scene representation without local-global fusion in terms of reconstruction and localization accuracy. Fig.~\ref{ic_loss_ab} presents the visualization of the reconstruction results. It can be observed that the complete model exhibits finer reconstructions and more reasonable predictions. The complete failure after removing the local representation is attributed to the difficulty of modeling large scenes using a smaller MLP and a lower number of iterations. Additionally, we employed direct concatenation for feature fusion. As shown in Tab.~\ref{ablation}, row e, and Fig.~\ref{ablation} (e), excessive padding is observed on the back of the chair.

\begin{table}[h]
\centering
\resizebox{\columnwidth}{!}{
\begin{tabular}{c|ccccc|cc}
\toprule
\multirow{2}{*}{Experiment}  & \multicolumn{5}{c|}{room0} & \multicolumn{2}{c}{fr3/walking-xyz}
\\&Acc. $\downarrow$ & Comp. $\downarrow$ & Comp.Ratio(\%) $\uparrow$ & ATE Mean $\downarrow$ & ATE RMSE $\downarrow$ & ATE Mean $\downarrow$ & ATE RMSE $\downarrow$ \\               
\midrule
\multicolumn{1}{l|}{a. Ours w/o local representation} & 90.45&84.31 &5.72 & 110.03&114.74 & 17.41&24.65 \\
\multicolumn{1}{l|}{b. Ours w/o global representation} &2.38 &1.85 & 97.17 &0.57 & 0.67 & 19.86& 24.93\\

\multicolumn{1}{l|}{c. Ours w/o attention-based feature fusion} &2.57&2.17&94.93 & 0.74& 0.89&18.09&22.53\\
\multicolumn{1}{l|}{d. Ours w/o result fusion} &2.37&1.80&97.02&0.49&0.60 &13.90&18.05\\
\multicolumn{1}{l|}{e. Ours w/ concatenation feature fusion }&2.38 & 1.80 & 97.10 & 0.63 &0.75 &16.85 &21.50\\
\multicolumn{1}{l|}{f. Full DVN-SLAM (Ours) } &\textbf{2.32} & \textbf{1.77}& \textbf{97.24} & \textbf{0.45} &\textbf{0.57} & \textbf{13.26}& \textbf{17.68} \\
\bottomrule
\end{tabular}}
\caption{Ablation study of local-global fusion neural implicit representation choices on room0 of Replica \cite{straub2019replica} and f3/walking-xyz of TUM\_RGBD \cite{sturm2012benchmark}. The results validate the effectiveness of each of our innovations.}
\label{ablation}
\end{table}

\begin{figure}[htbp]

\captionsetup{font=scriptsize}

\begin{floatrow}[2]
\figurebox{\caption{Visualization of information concentration loss ablation experiment.} \label{ic_loss_fig}}{%
  \includegraphics[width=\columnwidth]{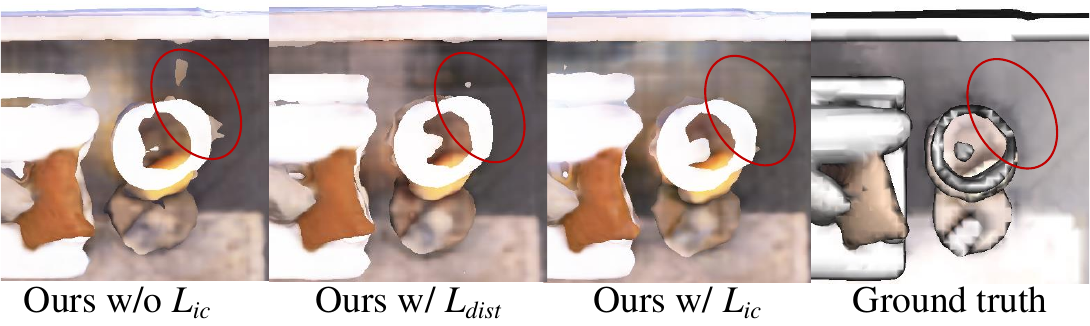}
  }%
\tablebox{}{%

\resizebox{\linewidth}{!}{
\begin{tabular}{c|cccc}
\toprule
\multicolumn{1}{l|}{Methods} & Acc. $\downarrow$ & Comp. $\downarrow$ & Comp.Ratio(\%) $\uparrow$ \\             
\midrule
\multicolumn{1}{l|}{Ours w/o $L_{ic}$} &2.38 &1.78 & 97.21\\
\multicolumn{1}{l|}{Ours w/ $L_{dist}$~\cite{barron2022mip}} &2.35 &1.78 & 97.23  \\
\multicolumn{1}{l|}{Ours w/ $L_{ic}$} & \textbf{2.32}& \textbf{1.77}& \textbf{97.24} \\
\bottomrule
\end{tabular}
\caption{Ablation study of information concentration loss.}
\label{ic_loss_tab}
}
}
\end{floatrow}
\end{figure}

\noindent \textbf{Effect of information concentration loss.}
Tab. \ref{ic_loss_tab} demonstrates the effect of information concentration loss removal. Fig. \ref{ic_loss_fig} illustrates the impact of information concentration loss on reconstruction, showcasing the beneficial optimization of scene information. It can suppress unnecessary reconstructions and improve reconstruction accuracy. In addition, we conducted a comparative experiment with deep regularization loss in Mip-NeRF 360~\cite{barron2022mip}.
The motivation behind depth regularization in Mip-NeRF 360~\cite{barron2022mip} is similar to our information concentration loss, which encourages each ray to be as compact as possible. The loss in Mip-NeRF 360~\cite{barron2022mip} ($L_{dist}$) consists of two terms: the first term minimizes the weighted distances between all pairs of interval midpoints, and the second term minimizes the weighted size of each individual interval. The computational complexity is $O(n^2)$. Our loss ($L_{ic}$) is obtained by the depth variance along the ray, encouraging that when sampled points are far from the depth center, their weights are as small as possible. The computational complexity is $O(n)$. The red circle in Fig. \ref{ic_loss_fig} shows the better suppression effect of our method for unnecessary reconstruction.

\section{Conclusion}
\label{sec:conclusion}

We propose DVN-SLAM, a real-time dynamic robust dense visual SLAM system based on a local-global fusion neural implicit representation. 
Thanks to the fusion representation and information concentration loss, DVN-SLAM is able to make reasonable predictions for the backside of objects, effectively avoiding holes and overfilling. More importantly, DVN-SLAM is dynamically robust and remains effective in high-dynamic scenes. Despite achieving promising results in indoor scenes, our proposed algorithm still faces challenges in applying it to larger-scale outdoor scenes. Challenges such as borderlessness, lighting variations, and sensor noise need to be addressed.

\bibliographystyle{unsrt}  
\bibliography{references} 

\appendix

\newpage
\begin{center}
\noindent {\large  Appendix}   
\end{center}

\section{Overview}
\label{sec: Overview}
In this document, we present more details and several
extra results as well as visualization. In Sec.~\ref{sec: dataset}, we
introduce details of the datasets used in our work. In Sec.~\ref{sec: implementation}, we elaborate on the implementation details of our method. The evaluation results for each scenario are presented in Sec.~\ref{sec: per-result}. Finally, we perform a performance analysis in Sec.~\ref{sec: performance}.

\section{Dataset Details}
\label{sec: dataset}
\subsection{Replica Dataset} The Replica dataset~\cite{straub2019replica} is a high-quality comprehensive scene dataset widely used in computer vision and machine learning research. It is jointly developed by Stanford University and the Princeton Visual AI Lab. The Replica dataset is based on realistic virtual environments and covers multiple scenarios, including apartments and offices. Each scene is carefully designed and modeled to represent different aspects and details of the real world. There are high-resolution RGB images, depth images, semantic annotations, and 3D reconstruction models for each scene. This paper primarily utilizes RGB images and depth maps to simulate the observations of a real RGB-D camera in the environment. The Replica dataset provides the pose of each frame as well as a dense, complete scene reconstruction mesh, which can serve as ground truth for localization and reconstruction, facilitating algorithm evaluation.

\subsection{TUM\_RGBD Dataset}
The TUM\_RGBD dataset~\cite{sturm2012benchmark} is another important dataset widely used in computer vision and machine learning research, provided by the Technical University of Munich (TUM) in Germany. This dataset is collected using RGB-D cameras in indoor environments and aims to provide rich visual and depth information for various computer vision tasks' research and evaluation. The TUM\_RGBD dataset consists of video sequences from multiple real-world scenes, including high-quality RGB images and corresponding depth maps. The dataset provides camera intrinsic and extrinsic parameters, as well as accurately measured camera poses, offering precise references for SLAM tasks. In addition to static scenes, the TUM\_RGBD dataset includes challenging dynamic scenes, allowing for the evaluation of our algorithms' dynamic robustness.

\subsection{Real-world Scene Data}

\begin{figure}[h] 
\center{\includegraphics[width=0.6\columnwidth]{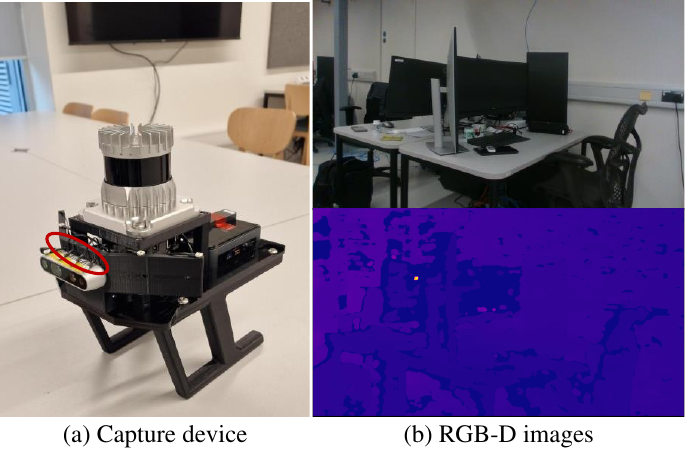} 
\caption{Capture device and captured RGB-D images. The RGB-D camera is the RealSense D435i. The data collection was performed in a laboratory setting.
}
\label{data}
}
\end{figure}

We collected real-world scene data using a handheld device as shown in Fig.~\ref{data} (a). The RGB-D camera utilized is the RealSense D435i. The small nodes positioned on top of the camera and the front wings are lasers, highlighted by red circles in the image, which are employed for the Phasespace motion capture localization system to obtain the ground truth trajectory. Fig.~\ref{data} (b) illustrates the RGB-D images captured in our experiments, with a resolution of $1280\times720$ and a frequency of 30 Hz. We capture a sequence of 6677 frames of images, along with the ground truth trajectory around a laboratory environment.

\section{Further Implementation Details}
\label{sec: implementation}
In this section, we provide additional detailed implementation information.
\subsection{Hyperparameters}

\noindent\textbf{Default setting.} 
In our method, for the global representation, we use 16 bins for OneBlob encoding~\cite{muller2019neural} in each dimension. Given coordinate $p= (x,y,z)$, this results in a 48-dimensional global feature. For the local representation, the resolution of the coarse and fine geometric feature planes $\{F_g^c, F_g^f\}$ is set to $\{24$ cm, $6$ cm$\}$. The resolution of the coarse and fine appearance feature planes $\{F_a^c, F_a^f \}$ is $\{24$cm, $3$ cm$\}$. The dimension of the feature stored in feature planes is 24. Both the  appearance and geometric feature decoders $\{\phi_a^1, \phi_a^2, \phi_g^1, \phi_g^2\}$ are two-layer MLPs. The input layer has a dimension of 48, and the hidden layer has a dimension of 16. The activation function for the intermediate layer is ReLU. The activation function for the output layer of the appearance decoders is Sigmoid, while the activation function for the output layer of the geometric decoders is Tanh. The weight $\omega$ for result fusion is set to 0.5. The truncation distance $tr$ is set to 6cm. Different loss weights are applied in the tracking and mapping processes. In the tracking process, the loss weights are set as $\lambda_{rgb} = 5, \lambda_{d} = 1, \lambda^n_{sdf} = 200, \lambda^f_{sdf} = 50,\lambda_{fs}=5,\lambda_{ic}=0.$ In the mapping process, the loss weights are set as $\lambda_{rgb} = 5, \lambda_{d} = 0.1, \lambda^n_{sdf} = 200, \lambda^f_{sdf} = 10,\lambda_{fs}=5, \lambda_{ic}=0.05$.

For rendering, we first sample $N_{strat}$ points for each ray through hierarchical sampling, followed by sampling $N_{imp}$ points near the surface. For pixels with ground truth depth, $N_{imp}$ additional points are uniformly sampled within the truncation distance around the depth measurement. As for pixels in depth discontinuities, we employ importance sampling~\cite{mildenhall2021nerf} based on the weights calculated from the stratified samples. Different sampling and iteration settings are used for different datasets.

\noindent\textbf{Replica dataset~\cite{straub2019replica}.} For the Replica dataset~\cite{straub2019replica}, $N_{strat}$ is set to 32, and $N_{imp}$ is set to 8. In the mapping process, we randomly select 4000 rays and perform 15 optimization iterations. In the tracking process, 2000 rays are randomly chosen, and 8 optimization iterations are performed.

\noindent \textbf{TUM\_RGBD dataset~\cite{sturm2012benchmark} and real-world scene data.} The setting for TUM\_RGBD dataset~\cite{sturm2012benchmark} and real-world scene data is the same. $N_{strat}$ is set to 48, and $N_{imp}$ is set to 8. In the mapping process, we randomly select 4000 rays and perform 30 optimization iterations. In the tracking process, 2000 rays are randomly chosen, and 30 optimization iterations are performed.

\subsection{Evaluation Metrics}

We follow the mesh culling strategy of Co-SLAM~\cite{wang2023co} to process the reconstructed mesh, removing any unobserved regions outside the camera frustum and noise points inside the camera frustum but outside the target scene. We evaluate the reconstruction quality using \textbf{Depth L1} (cm), \textbf{Acc}uracy (cm), \textbf{Comp}letion (cm), and \textbf{Comp}letion \textbf{Ratio} (\%) thresholds set to $5$ cm. For tracking evaluation, we utilize the Average Translation Error (ATE) mean and Root Mean Square Error (RMSE) in centimeters.

For Depth L1, we generate depth maps from the reconstructed mesh and ground truth mesh at 1000 viewpoints. Depth L1 is defined as the average absolute error between the reconstructed and ground truth depth maps. For Accuracy, Completion, and Completion Ratio, we sample 200,000 points from the reconstructed mesh and the ground truth mesh. Accuracy is defined as the average distance between sampled points in the reconstructed mesh and their nearest ground truth points. Completion is defined as the average distance between sampled points from the ground-truth mesh and the nearest reconstructed. Completion Ratio is defined as the percentage of points in the reconstructed mesh with completion under $5$ cm. The formulas for these metrics are as follows:
\begin{equation}
Depth L1 = \frac{1}{1000H\times W}  \sum |\hat{d} -d_{gt}|,
\end{equation}
\begin{equation}
Acc = \frac{1}{|P|}  \sum_{p \in P} \min_{q \in Q }||p-q||,
\end{equation}
\begin{equation}
Comp = \frac{1}{|Q|}  \sum_{q \in Q} \min_{p \in P} ||p-q||,
\end{equation}
\begin{equation}
Comp\: Ratio = \frac{1}{|Q|}  \sum_{q \in Q} \min_{p \in P} ||p-q|| < 5,
\end{equation}
where $H$ and $W$ represent the resolution of the depth maps. $\hat{d}$ and $d_{gt}$ correspond to the depth maps generated from the reconstructed mesh and the ground truth mesh, respectively. $P$ and $Q$ represent point clouds sampled from the reconstructed mesh and the ground truth mesh, respectively.

\section{Per-Scene Breakdown of the Results}
\label{sec: per-result}
In this section, we present more detailed evaluation results. Tab.~\ref{replica_all} displays the per-scene quantitative evaluation results of our method and others on the Replica dataset. It can be observed that our method achieves the best performance in Depth L1, ATE mean, and ATE RMSE across all scenes. Competitive results are also obtained in other metrics.

\begin{figure*}[t] 
\center{\includegraphics[width=1.0\textwidth]{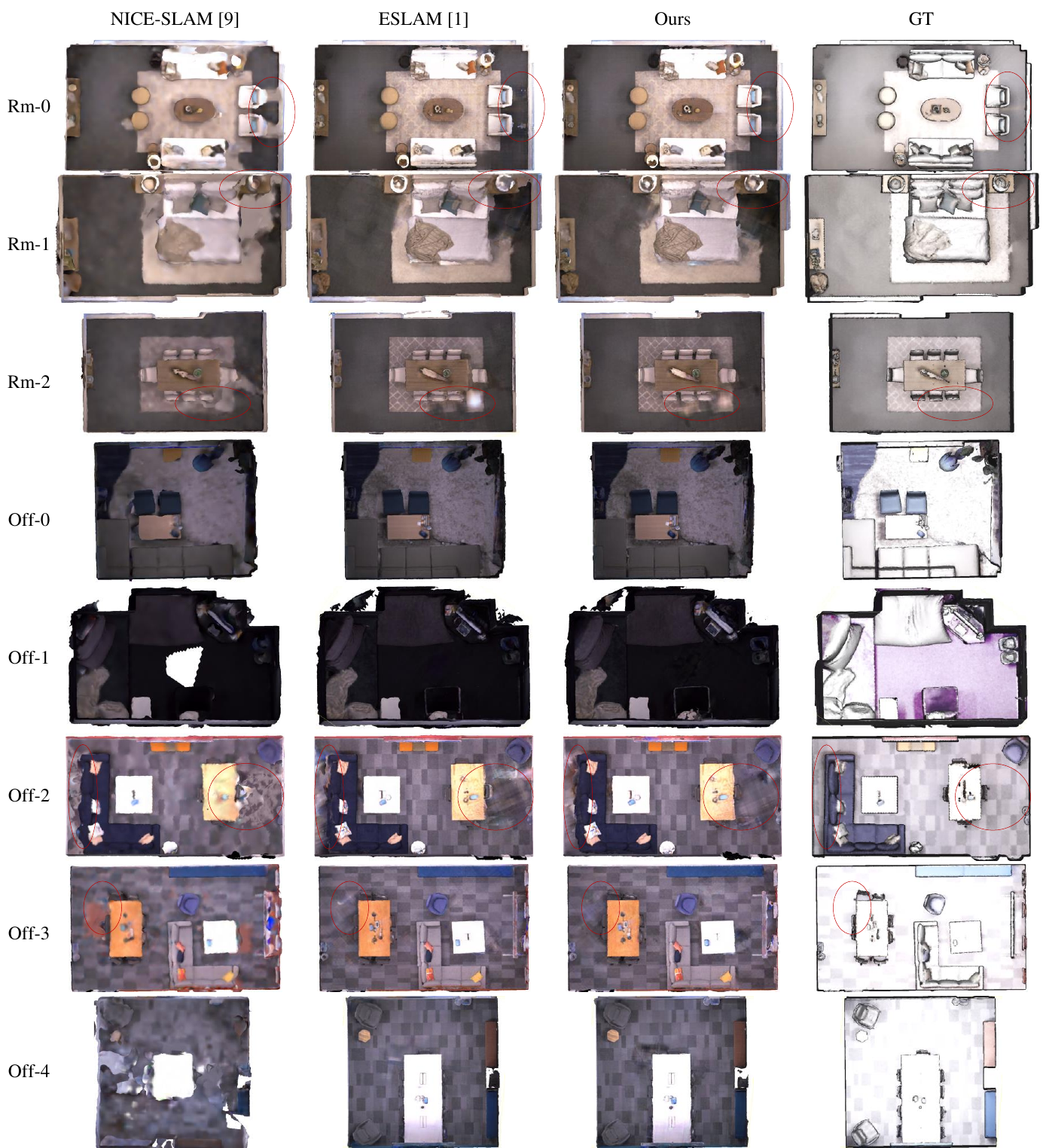} 
\caption{Reconstruction results of the Replica dataset~\cite{straub2019replica}. The highlighted red circle emphasizes the remarkable modeling capability of our method for inadequately observed regions. It can be observed that our method demonstrates finer modeling of local details and a more complete representation of global structures.
}
\label{replica_mesh}
}
\end{figure*}

Fig.~\ref{replica_mesh} visualizes the reconstruction results for each scene in the Replica dataset~\cite{straub2019replica}. We highlight areas where our method exhibits significantly superior performance with red circles. It can be observed that our method demonstrates finer modeling of local details and a more complete representation of global structures. In regions with insufficient observations, our method excels in predicting and supplementing the reconstruction.

\begin{figure*}[t] 
\center{\includegraphics[width=1.0\textwidth]{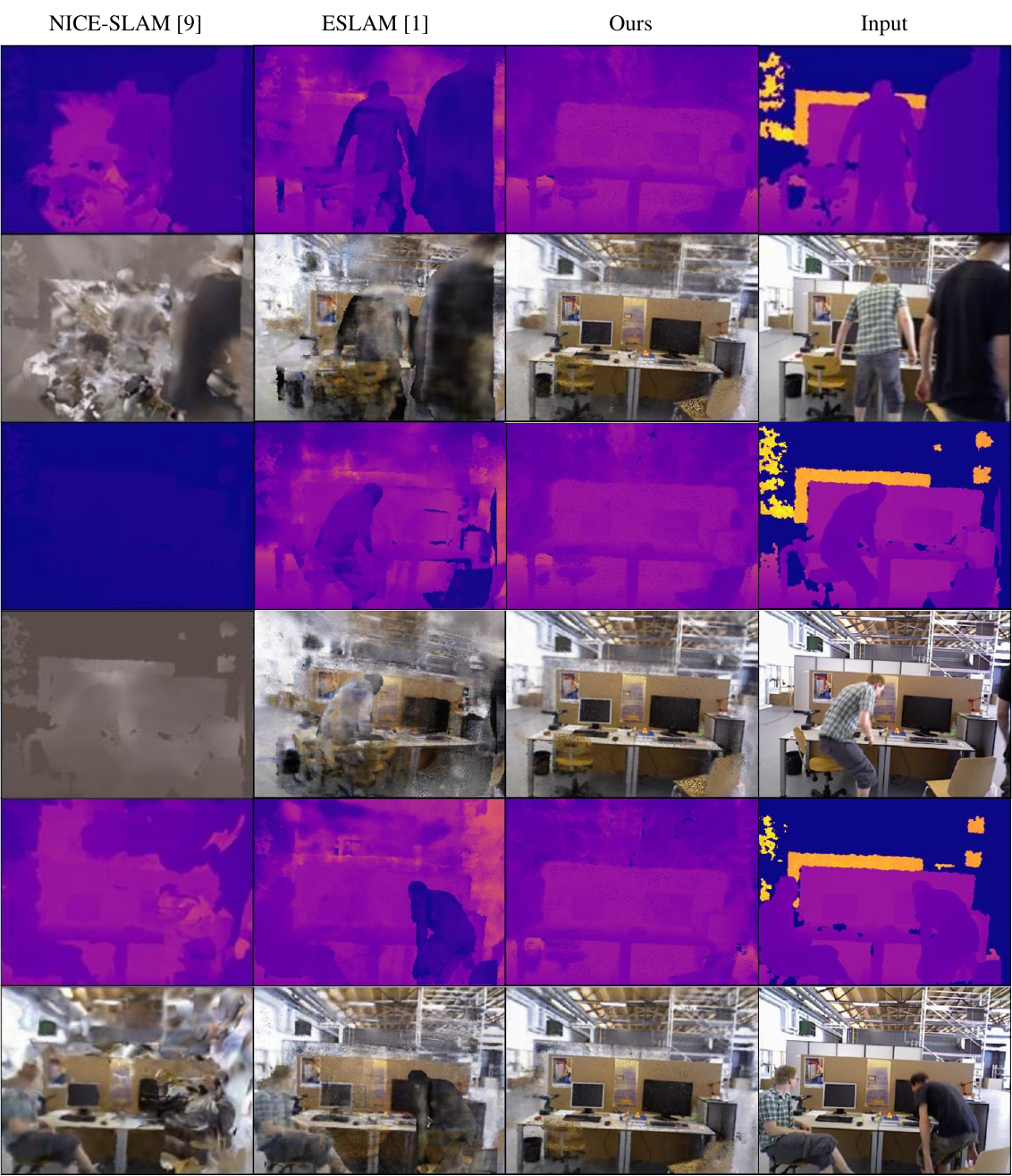} 
\caption{Rendering results in dynamic scenes of the TUM\_RGBD dataset~\cite{sturm2012benchmark}. Our method not only produces clearer rendering in static regions but also effectively filters out dynamic objects and fills in the corresponding background. In contrast, other methods produce blurry renderings and retain dynamic occlusions.
}
\label{tum_render}
}
\end{figure*}

Fig.~\ref{tum_render} illustrates the rendering visualization in dynamic scenes of the TUM
-RGBD dataset~\cite{sturm2012benchmark}. Our method produces clearer rendering results, and it effectively ignores dynamic objects while filling in occluded backgrounds. 

\section{Performance Analysis}
\label{sec: performance}

\begin{table}[h]
\centering
\resizebox{0.7\textwidth}{!}{
\begin{tabular}{c|ccc}
\toprule
Method & Tracking FPS $\uparrow$ & Mapping FPS $\uparrow$ & Param $\downarrow$\\   
\midrule
iMAP*~\cite{sucar2021imap} &9.92 &2.23 &\textbf{0.26M}\\
NICE-SLAM~\cite{zhu2022nice}& 13.70&0.20 &17.4M\\
Vox-Fusion~\cite{yang2022vox}&2.11 &2.17 & \underline{0.87M} \\
Co-SLAM~\cite{wang2023co}  & \underline{17.24}& \textbf{10.20}&\textbf{0.26M}\\
ESLAM~\cite{johari2023eslam} &\textbf{18.11} &\underline{3.62} &9.29M\\
DLGF-SLAM (Ours)&9.62 & 1.82 & 7.22M\\

\bottomrule
\end{tabular}
}
\caption{Runtime and memory. Our method boasts competitive real-time performance and memory usage.}
\label{runtime}
\end{table}

\textbf{Runtime and memory.}
Tab \ref{runtime} illustrates a comparison of the runtime and memory between our algorithm and other methods. Our method achieves 7.32 Hz tracking and 1.67 Hz mapping performance on a server equipped with an NVIDIA A100 GPU. Due to the more complex scene representation, our algorithm exhibits slightly lower real-time performance compared to ESLAM~\cite{johari2023eslam} and Co-SLAM~\cite{wang2023co}. In comparison to ESLAM~\cite{johari2023eslam}, which also utilizes feature planes, we use a smaller feature dimension, resulting in lower memory requirements. Improving real-time performance and reducing memory usage further will be considered in our future work.

\begin{table*}
    \centering
    \small
    \resizebox{\columnwidth}{!}{
    \begin{tabular}{c|c|cccc|cc}
    \toprule
    & \multirow{2}{*}{Methods} & \multicolumn{4}{c|}{Reconstruction [cm]} & \multicolumn{2}{c}{Localization [cm]}    \\
    && Depth L1 $\downarrow$ & Acc. $\downarrow$ & Comp. $\downarrow$ & Comp.Ratio(\%) $\uparrow$ & ATE Mean $\downarrow$ & ATE RMSE $\downarrow$ \\          
    \midrule
    \multirow{6}{*}{Rm-0}                     
    &iMAP*~\cite{sucar2021imap}  & 5.08 & 4.01 & 5.84 & 78.34 & 3.12 & 5.23 \\
    &NICE-SLAM~\cite{zhu2022nice} & 1.79 & 2.44 & 2.60 & 91.81 & 1.43 & 1.69\\
    &Vox-Fusion~\cite{yang2022vox} & 1.76 & \textbf{1.77} & 2.69 & 92.03 &1.03  &1.37 \\
    &Co-SLAM~\cite{wang2023co} & 1.05 & \underline{2.11} & \underline{2.02} & 95.26 & 0.73  &0.84 \\
    &ESLAM~\cite{johari2023eslam} & \underline{0.73} & 2.45 & \textbf{1.79} & \textbf{97.29} & \underline{0.61} &\underline{0.71} \\
    &DLGF-SLAM (Ours) & \textbf{0.60} & 2.32 & \textbf{1.77} &\underline{97.24}  &\textbf{0.45}  &\textbf{0.57}\\    

    \midrule
    \multirow{6}{*}{Rm-1}  
    &iMAP*~\cite{sucar2021imap}  & 3.44 & 3.04 & 4.40 & 85.85 & 2.54 &3.09  \\
    &NICE-SLAM~\cite{zhu2022nice} & 1.33 & 2.10 & 2.19 & 93.56 &1.70  &2.13 \\
    &Vox-Fusion~\cite{yang2022vox} & 2.52 & \textbf{1.51} & 2.31 & 92.47 & 1.35 & 1.9\\
    &Co-SLAM~\cite{wang2023co} & 0.85 & \underline{1.68} & \underline{1.81}	& 95.19 & 0.88 & 1.32\\
    &ESLAM~\cite{johari2023eslam} & \underline{0.74} & 2.44 & \textbf{1.58} & \textbf{96.80} &  \underline{0.56}&\underline{0.70} \\
    &DLGF-SLAM (Ours) & \textbf{0.57} &1.89& \textbf{1.58} &\underline{96.41}  &\textbf{0.52}  &\textbf{0.62}\\

    \midrule
    \multirow{6}{*}{Rm-2}     
    &iMAP*~\cite{sucar2021imap}  & 5.78 & 3.84 & 5.07 & 79.40 & 2.31 &2.58  \\
    &NICE-SLAM~\cite{zhu2022nice} & 2.20 & 2.17 & 2.73 & 91.48 & 1.41 & 1.87\\
    &Vox-Fusion~\cite{yang2022vox} & 3.58 & 2.23 & 2.58 & 90.13 & 1.02 &1.47 \\
    &Co-SLAM~\cite{wang2023co} & 2.37 & 1.99 & 1.96 & 93.58 & 0.93 & 1.22\\
    &ESLAM~\cite{johari2023eslam} & \underline{1.26} & \underline{1.70} &\underline{1.61}  & \underline{96.89} &\underline{0.43}  &\underline{0.52} \\
    &DLGF-SLAM (Ours) &\textbf{0.90}  &\textbf{1.63}  &\textbf{1.60}  &\textbf{97.06}  & \textbf{0.38} &\textbf{0.47}\\  
    \midrule
    \multirow{6}{*}{Off-0}                  
    &iMAP*~\cite{sucar2021imap} & 3.79 & 3.34 & 3.62 & 83.59 & 1.69 & 2.40 \\
    &NICE-SLAM~\cite{zhu2022nice} & 1.43 & 1.85 & 1.84 & 94.93 & 1.12 &1.26 \\
    &Vox-Fusion~\cite{yang2022vox} & 3.44 & 1.63 & 1.87 & 93.86 &0.98  &1.35 \\
    &Co-SLAM~\cite{wang2023co} & 1.24 & \underline{1.57} & \underline{1.56} & 96.09 & 0.53 & 0.64\\
    &ESLAM~\cite{johari2023eslam} & \underline{0.71} & \textbf{1.48} &\textbf{1.30}  &\underline{98.45}  &\underline{0.42}  &\underline{0.57} \\
    &DLGF-SLAM (Ours)  & \textbf{0.54} &1.54  &\textbf{1.30}  & \textbf{98.64} &\textbf{0.40}  &\textbf{0.45}\\
    
    \midrule
    \multirow{6}{*}{Off-1}                       
    &iMAP*~\cite{sucar2021imap}  & 3.76 & 2.10 & 3.62 & 88.45 & 1.03 &1.17  \\
    &NICE-SLAM~\cite{zhu2022nice} & 1.58 & 1.56 & 1.82 & 94.11 & 0.74 &0.84 \\
    &Vox-Fusion~\cite{yang2022vox} & 1.77 & \underline{1.44} & 1.66 & 94.40 & 1.29 & 1.76\\
    &Co-SLAM~\cite{wang2023co} & 1.48 & \textbf{1.31} & 1.59 & 94.65 & 0.48 & \underline{0.54}\\
    &ESLAM~\cite{johari2023eslam} & \underline{1.02} & 1.60 &\underline{1.47}  &\underline{96.04}  &  \underline{0.46}& 0.55\\
    &DLGF-SLAM (Ours) & \textbf{0.99} & 1.81 & \textbf{1.24} & \textbf{97.64} & \textbf{0.36} &\textbf{0.45}\\

    \midrule
    \multirow{6}{*}{Off-2}                  
    &iMAP*~\cite{sucar2021imap}  & 3.97 & 4.06 & 4.73 & 79.73 & 3.99 &5.67  \\
    &NICE-SLAM~\cite{zhu2022nice} & 2.70 & 3.28 & 3.11 & 88.27 & 1.42 & 1.71\\
    &Vox-Fusion~\cite{yang2022vox} & 3.52 & \textbf{2.09} & 3.03 & 88.94 &0.73  &1.18 \\
    &Co-SLAM~\cite{wang2023co} & 1.86 & 2.84 & 2.43 & 91.63 & 1.88 &2.07\\
    &ESLAM~\cite{johari2023eslam} & \underline{0.93} & \underline{2.55} & \underline{2.05} &\textbf{96.14}  & \textbf{0.47} &\textbf{0.58} \\
    &DLGF-SLAM (Ours) & \textbf{0.89} & 2.93 & \textbf{1.88} &\underline{95.27}  &\underline{0.55}  &\underline{0.63}\\

    \midrule
    \multirow{6}{*}{Off-3}                    
    &iMAP*~\cite{sucar2021imap}  & 5.61 & 4.20 & 5.49 & 73.90 & 4.05 &5.08  \\
    &NICE-SLAM~\cite{zhu2022nice} & 2.10 & 3.01 & 3.16 & 87.68 &  2.31&3.98 \\
    &Vox-Fusion~\cite{yang2022vox} & 1.82 & \textbf{2.33} & 2.81 & 89.10 &0.69  &1.11 \\
    &Co-SLAM~\cite{wang2023co} & 1.66 & 3.06 & 2.72 & 90.72 & 1.36 & 1.43\\
    &ESLAM~\cite{johari2023eslam} & \underline{1.03} &  \underline{2.38}&\underline{2.18}  &\textbf{95.33}  &\underline{0.61}  &\underline{0.72} \\
    &DLGF-SLAM (Ours) & \textbf{0.73} & 2.52 &  \textbf{2.07}& \underline{95.17} &\textbf{0.45}  &\textbf{0.53}\\

    \midrule
    \multirow{6}{*}{Off-4}
    &iMAP*~\cite{sucar2021imap}  & 5.71 & 4.34 & 6.65 & 74.77 & 1.93 &2.23  \\
    &NICE-SLAM~\cite{zhu2022nice} & 2.06 & 2.54 & 3.61 & 87.23 & 2.22 &2.82 \\
    &Vox-Fusion~\cite{yang2022vox} & 4.84 & \textbf{2.02} & 3.51 & 86.53 &1.18  &1.64 \\
    &Co-SLAM~\cite{wang2023co} & 1.54 & 2.23 & 2.52 & 90.44 & 0.74 &0.87 \\
    &ESLAM~\cite{johari2023eslam} & \underline{1.18} & 2.06 & \textbf{2.05} &\textbf{94.53}  & \underline{0.52} &\underline{0.63} \\
    &DLGF-SLAM (Ours) & \textbf{0.81} &\underline{2.04}  &\underline{2.11}  &\underline{94.04}  &\textbf{0.44}  &\textbf{0.51} \\

    \bottomrule
    \end{tabular}}
    \caption{Quantitative comparison of DLGF-SLAM with existing nerf-based dense V-SLAM in terms of reconstruction and localization accuracy on the Replica dataset \cite{straub2019replica}. 
    The results are reported as the average over 5 runs on each scene of the Replica dataset \cite{straub2019replica}. Our approach demonstrates significant improvements over previous works.  }
    \label{replica_all}
\end{table*}

\end{document}